\documentclass[10pt,twocolumn,letterpaper]{article}

\usepackage{cvpr}              % To produce the CAMERA-READY version
\usepackage[dvipsnames]{xcolor}

\newcommand{\ts}[1]{\textsuperscript{#1}}
\usepackage{graphicx}
\usepackage{amsmath}
\usepackage{amssymb}
\usepackage{booktabs}
\usepackage{amsfonts}

\usepackage{float} % For positioning tables
\usepackage{array}
\usepackage{geometry}
\usepackage{fontawesome}
\usepackage{adjustbox}
\usepackage{tabularx}
\newcommand{\quoteArg}[1]{``#1''}

\newcommand{\Frst}[1]{\textcolor{red}{\textbf{#1}}}

\newcommand{\Blue}[1]{\textcolor{Blue}{\textbf{#1}}}
\newcommand{\Brown}[1]{\textcolor{Brown}{\textbf{#1}}}
\newcolumntype{P}[1]{>{\raggedright\arraybackslash}p{#1}}

\definecolor{cvprblue}{rgb}{0.21,0.49,0.74}
\usepackage[pagebackref,breaklinks,colorlinks,citecolor=cvprblue]{hyperref}

\def\paperID{6} % *** Enter the Paper ID here
\def\confName{CVPR}
\def\confYear{2024}

\title{Towards Automating the Retrospective Generation of BIM Models: A Unified Framework for 3D Semantic Reconstruction of the Built Environment}

\author{Ka Lung Cheung\ts{1,2} \quad Chi~Chung~Lee\ts{2}\\ 
\ts{1}The Chinese University of Hong Kong \quad \ts{2}Hong Kong Metropolitan University\\
{\tt \small klcheung@mae.cuhk.edu.hk, cclee@hkmu.edu.hk}\\
\small \url{https://github.com/Semanticity-Research/SRBIM}}

\begin{document}
\maketitle
% \begin{abstract}
% The adoption of Building Information Modeling (BIM) is beneficial in construction projects. However, it faces challenges due to the lack of a unified, efficient, and scalable framework for converting 3D model details into BIM. This paper introduces SRBIM, a unified semantic reconstruction architecture for BIM generation. SRBIM comprises streamlined processes: 1) Semantic segmentation of built components with PTv2; 2) Mesh reconstruction, refinement to generate a semantically-enriched mesh segment with our proposed MFS module; 3) IFC object mapping pipeline followed by partition-based colorization to reconstruct high-fidelity, visually appealing BIM models. Our approach's effectiveness is demonstrated through extensive qualitative and quantitative evaluations, establishing a new paradigm for automated BIM modeling. 

% \end{abstract}

% Se
\begin{abstract}
The adoption of Building Information Modeling (BIM) is beneficial in construction projects. However, it faces challenges due to the lack of a unified and scalable framework for converting 3D model details into BIM. This paper introduces SRBIM, a unified semantic reconstruction architecture for BIM generation. Our approach's effectiveness is demonstrated through extensive qualitative and quantitative evaluations, establishing a new paradigm for automated BIM modeling.
% Further testing results and our segmentation dataset for 3D point cloud models are available at \url{https://github.com/BIMSRLab/SRBIM}.

\end{abstract}
% Second, we introduce the first semantically-enriched, publicly available 3D building model dataset for BIM reconstruction and 3D semantic segmentation, featuring real-world buildings and an open architectural landscape in Hong Kong.
%%%%%%%%% BODY TEXT
\section{Introduction}
% New intro 15022024

% \begin{figure}
%     \centering
%     % \includegraphics[width=1\linewidth]{img/HK
%     \includegraphics[width=1\linewidth]{img/HKBIM_dataset_250.pdf}
%     \caption{3D Architectural models from our SRBIM dataset. Different semantic classes are labeled by different colors. We present a dataset of 3D building models annotated for prominent buildings and landscapes in Hong Kong. Additionally, we introduce a semantic reconstruction pipeline for BIM that streamlines 3D semantic segmentation, mesh reconstruction, refinement, and the reconstruction of high-fidelity BIM models.}
%     \label{Fig:SRBIM Dataset}
% \end{figure}

% \begin{figure}
%     \centering
%     % \includegraphics[width=1\linewidth]{img/HK
%     \includegraphics[width=1\linewidth]{img/HKBIM_dataset_1_3.png}
%     \caption{We present a dataset of 3D models annotated for prominent buildings and landscapes in Hong Kong. Additionally, we introduce a semantic reconstruction pipeline for Built Information Modeling (BIM) that streamlines 3D semantic segmentation, mesh reconstruction, refinement, and the reconstruction of high-fidelity BIM models.}
%     \label{Fig:SRBIM Dataset}
% \end{figure}

The construction sector has seen a surge in the adoption of Building Information Modeling (BIM),  which enhances collaboration and transparency \cite{schonfelder_automating_2023,volk_building_2014,wiley_bim_2024,croce_semantic_2021,gu_understanding_2010}. However, this process can be error-prone and time-consuming, often requiring specialized knowledge. Moreover, there is also a shortage \cite{schonfelder_automating_2023,mishra_holistic_2024,ullah_overview_2019,shin_facilitators_2021} of BIM experts in many countries , impede the widespread adoption of BIM. 

Automated BIM generation with machine-learning techniques offers promising solutions to alleviate related issues.\cite{schonfelder_automating_2023, mishra_holistic_2024, shin_facilitators_2021}. Nonetheless, an open challenge still remains for a unified framework capable of converting spatial and semantic information from a 3D model into BIM \cite{kang_rule-based_2020}. The challenge is mainly induced by the complexity of the BIM generation process, which involves distinguishing architectural elements, colorization, creating visual representations, and assembling these components into a semantically-enriched BIM. Despite the proposition of several automated BIM modeling frameworks, most primarily focus on reconstructing the 3D geometry of high-rise buildings \cite{bassier_classification_2019, lee_recognizing_2019, ma_semantic_2020, ferrari_neural_2018} and extracting the semantic information of indoor building components \cite{zhang_transformer-based_2023,ma_semantic_2020,hu_robot-assisted_2023}. Yet, methods for outdoor facade-level or infrastructure projects remain largely unexplored due to the lack of data sources, robust machine learning methods, or a streamlined approach \cite{schonfelder_automating_2023}.

% However, there remains a research gap for a unified, automated solution that can convert spatial and semantic information from a 3D model into BIM \cite{kang_rule-based_2020}, considering the number of steps involved in BIM generation (\eg distinguishing architectural elements, colorization, visual representation creation, and assembling these components into a semantically-enriched BIM); Several automated BIM modeling framework were proposed to reconstruct the 3D geometry of high-rise buildings\cite{bassier_classification_2019,lee_recognizing_2019,ma_semantic_2020,ferrari_neural_2018} or extracting the semantic information of indoor building components \cite{zhang_transformer-based_2023}, but methods for outdoor fa\c{c}ade-level or infrastructure projects are remainly widely unexplored due to the lack of data sources ,robust ML methods and unified pipeline.\cite{schonfelder_automating_2023,mishra_holistic_2024,ullah_overview_2019,shin_facilitators_2021} of BIM experts.

% The \textbf{second} issue arises from current methods for creating digital building models, which often overlook outdoor fa\c{c}ade-level or infrastructure projects \cite{schonfelder_automating_2023}. Most of these methods mainly focus on reconstructing the 3D geometry of high-rise buildings\cite{bassier_classification_2019,lee_recognizing_2019,ma_semantic_2020,ferrari_neural_2018}. with few research methods exist for extracting the semantic information of building components \cite{zhang_transformer-based_2023}.

In this paper, we present a novel \textbf{S}emantic \textbf{R}econstruction framework for universal-scale 3D constructed space to automate the \textbf{BIM} documentation (\textbf{SRBIM}). In contrast to previous studies \cite{bassier_classification_2019,zhuo_facade_2019,obrock_first_2018,grilli_machine_2020,pierdicca_point_2020,lee_recognizing_2019,ma_semantic_2020,matrone_transfer_2021,nivaggioli_using_2019,yin_automated_2021,ferrari_neural_2018,barazzetti_vault_2019,balado_road_2019,barrile_road_2020,lee_semantic_2021,park_semantic_2021,narazaki_visionbased_2020,yang_automated_2022}, which focus solely on one or a few of BIM generation process. SRBIM comprises streamlined processes: 1) Semantic segmentation of built components with Point Transformer V2 (PTv2) \cite{wu_point_2022}; 2) Mesh reconstruction, refinement to generate a semantically-enriched mesh segment with our proposed MFS module; 3) Industrial Foundation Class (IFC) \cite{noauthor_industry_2019} object mapping pipeline followed by partition-based colorization to reconstruct semantically-enriched, visually appealing BIM models.

\section{Architecture}
\begin{figure*}[t]
    \centering
    \includegraphics[width=0.8\linewidth]{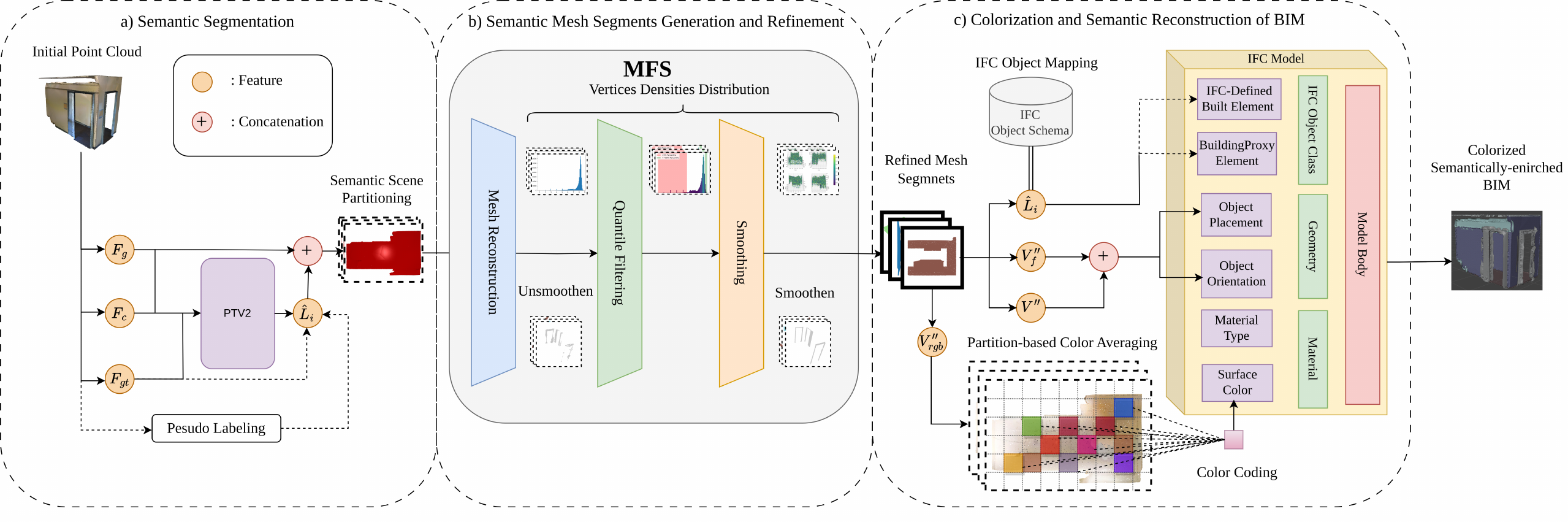}
    \caption{Overview of the SRBIM pipeline.
   In the first step (a), the transformer model, PTv2, combines both spatial $F_{g}$ and color $F_{c}$ features to predict labels $\hat{L}i$. The model uses ground truth data $F_{gt}$ for supervised learning or as a substitute of $\hat{L}i$, with pseudo-labeling acting as an alternative to replacing $\hat{L}i$ when $F_{gt}$ is missing. This is followed by steps (b) and (c), where the initial point cloud scene is digitized into BIM.}
   \label{Fig:SRBIM-pipeline-overview}
\end{figure*}

SRBIM (\cf \cref{Fig:SRBIM-pipeline-overview}) is built on three core components: trained segmentation model PTv2, MFS module, and the BIM reconstruction pipeline.
% The segmentation model extracts semantic information from the input point cloud scene. The MFS module then transforms these semantically-enriched point segments into refined mesh segments. Lastly, the BIM reconstruction pipeline reassembles into a semantically-enriched BIM, which contains the visual and material representations of the built components.

% \begin{figure}
%     \centering
%     \includegraphics[width=1\linewidth]{add_view_SRBIM_model.drawio.png}
%     \caption{Enter Caption}
%     \label{fig:enter-label}
% \end{figure}

% \begin{figure}
%     \centering
%     \includegraphics[width=0.5\linewidth]{add_view_SRBIM_model.drawio.png}
%     \caption{Enter Caption}
%     \label{fig:enter-label}
% \end{figure}

% \begin{figure}
%     \centering
%     % \includegraphics[width=1\linewidth]{img/SRBIM_step1-3.png}
%     \includegraphics[width=1\linewidth]{img/SRBIM-2602.pdf}
%     \caption{Overview of the SRBIM pipeline.
%    In the first step (a), the transformer model, PTv2, combines both spatial $F_{g}$ and contextual $F_{c}$ features to predict labels $\hat{L}i$. The model uses ground truth data $F_{gt}$ for supervised learning or as a substitute of $\hat{L}i$, with pseudo labeling acting as an alternative to replacing $\hat{L}i$ when $F_{gt}$ is missing. This is followed by steps (b) and (c), where the initial point cloud scene is digitized into BIM.}
%    \label{Fig:SRBIM-pipeline-overview}
% \end{figure}

\subsection{Semantic Segmentation with PTv2}
Considering the great success of ViT in semantic segmentation \cite{lu_3dctn_2022,guo_pct_2021,yu_3d_2021,xu_tdnet_2022}, PTv2 \cite{wu_point_2022} is proposed for point-wise semantic segmentation in our study. For the initial point cloud features, we first embed the input channels with a basic block with an attention group. With the encoding and decoding stage, PTv2 output head remap the final logits $\hat{L}_i \in \mathbb{R} $ (\ie the class label) for each point \(p_i\) and achieves the scene classification for each scene $S = (F_g, F_c)$, $S$ be a 3D point cloud scene containing a set of points \(p_i = (f_g, f_c) \in S\), where \(f_g \in \mathbb{R}^3\) and \(f_c \in \mathbb{R}^3\) represents the point position and color features (\ie RGB values), respectively.
Given \(f_c\) contains information related to \(F_{gt}\) for supervised training with PTv2, and considering $F_{gt} =\{f_{gt_1},f_{gt_2},\ldots,f_{gt_n}\}$, where $n$ represents the total number of individual ground truth labels within the set, the relationship can be described as \(f_c \supseteq \{f_{gt}\} \in F_{gt}\).

% \subsubsection{Pseudo-labeling.}In the process of semantic segmentation of point cloud for the built environment, ground truth labels \( F_{gt} \) are essential for supervised learning. However, there are scenarios in the absence of ground truth labels $F_{gt} = \emptyset$, and for specific scenarios, the goal is not to distinguish between different built object classes but to generate the initial labels based on some criteria (\eg initial labels are continuously updated with the topology, spatial relationship, eigenvector, singular value, geometric features or density). In such cases,  every point is randomly assigned a single semantic class label from the set of all possible semantic class labels \(L_{\text{class}} = \{L_1, L_2, \ldots, L_k\}\). It is arranged so that each point is assigned a label within the same class \(\hat{L}_i= L_i \in \mathbb{R}\), where \(k\) represents the total number of unique semantic classes. Consider that pseudo-labeling is a time \(O(1)\) operation; the time and space complexity remains constant regardless of the size of the input. This unified class assignment treats the point cloud as a single, undifferentiated entity, expecting a single segment in the partitioning step.
\paragraph{Semantic Scene Partitioning.}
The partitioning divides semantically-enriched point cloud scene $\tilde{S} = \{(f_g, f_c, \hat{L}_i) \mid (f_g, f_c) \in S, \, \hat{L}_i \in \mathbb{R}\}$ into point segments based on semantics classes $L_{\text{class}}$, organizing the cloud into a set of segments \( \mathcal{C} = \{C_1, C_2, \ldots, C_q\} \). Each segment \(C_i\), for \(i\) ranging from 1 to \(q\), groups points that share the same semantic label $\hat{L}_i$, streamlining the cloud into \(q\) distinct, semantically-enriched segments.

\subsection{Mesh Segments Generation and Refinement}
% \textbf{Mesh Segments Generation and Refinement.}
We introduce the \textbf{MFS} module that sequentially executes \textbf{M}esh reconstruction, quantile \textbf{F}iltering of outlier mesh per-vertex, and final mesh \textbf{S}moothing. These steps are performed for each pre-labeled point cloud partition, converting them into mesh segments.

\paragraph{Mesh Reconstruction.}
Poisson Surface Reconstruction \cite{kazhdan_poisson_2006} is adopted in creating initial meshes $ \mathcal{M} = \{m_1, m_2, \ldots, m_q\} $ as it can preserve sharp features and details for intricate indoor and outdoor 3D scans. 

% Its gradient-based formulation enables it to retain high fidelity for complex 3D scenes. 
% This is achieved by setting the search depth to 9 and the number of iterations to 10. These parameters control the detail of the resulting mesh segments. 
% Once set, the point segments are fed into the Poisson reconstruction pipeline.
\paragraph{Quantile Filtering of Outlier Mesh Vertices.} We compute and normalize the densities of each vertex $v$ in the newly generated mesh. The densities are uniformly adjusted across all vertices, expressed as
\(\tilde{\delta}_i = \frac{\delta_i}{\max_{1 \leq j \leq n} (\delta_j)}
\), which normalizes them between 0 and 1. A threshold quantile (\(\alpha = 0.05\)) is then set to identify vertices with low density. Vertices with a normalized density below \(\alpha\) are recognized as \(V_{\text{remove}} = \{ v_i | \tilde{\delta_i} < \alpha \}\). The interim mesh \(m\) is updated by removing \(V_{\text{remove}}\), producing a refined mesh \(m'\).
\paragraph{Mesh Smoothing.}
Laplacian Smoothing \cite{herrmann_laplacian-isoparametric_1976} is applied to updated segment \(m'\) by adjusting the position of $v'_i$ based on the average of their neighboring vertices. Applying this method smooths out irregularities, resulting in mesh segments \( \mathcal{M''} = \{m''_1, m''_2, \ldots, m''_k\} \), reflects the completed mesh shapes of the point cloud scene $S$.

\subsection{BIM Reconstruction}
% \textbf{BIM Reconstruction}
% % \subsubsection{Initializing Project.}
% % The reconstruction process begins with the creation of the \quoteArg{IfcProject} entity and a context identifier. This process sets up the project structure, including unique units of measurement and 3D space. Material class, transparency, and material settings are also initialized to provide a color-coded surface template.

\paragraph{Mapping with IFC Schema.} To provide semantic information about the built environment, the refined, semantically-enriched mesh segments $\mathcal{M''}$ are converted into IFC class objects. Each segment $m''$ is registered and assigned a unique context identifier. If the label $\hat{L}_i$ ends with the same name as one of the IFC class objects in the schema, it is identified as a specific IFC-predefined element. Segments that do not find matches with the IFC schema are mapped as \quoteArg{IfcBuildingElementProxy}- a class that shares common information among many instances of the same proxy without establishing a specific semantic meaning of the type. 
% The mapping implicit function is defined as :
% % \begin{enumerate}[itemsep=0pt,topsep=2bp]
%         \begin{align}
%             \label{eq:eq1}
%                 f(m) =
%               \begin{cases} 
%                \text{IFC class}(m'') \quad \text{if } \exists \, \text{IFC class for } m'' \\
%                \text{IfcBuildingElementProxy}(m'') \quad \text{otherwise}
%               \end{cases}
%         \end{align}
% % \end{enumerate}
% % \begin{equation}
% % f(m) =
% %   \begin{cases} 
% %    \text{IFC class}(m'') & \text{if } \exists \, \text{IFC class for } m'' \\
% %    \text{IfcBuildingElementProxy}(m'') & \text{otherwise}
% %   \end{cases}
% % \end{equation}

% % \begin{equation}
% % f(m) =
% %   \begin{cases} 
% %    \text{IFC class}(m'') & \text{if } \exists \, \text{IFC class for } m'' \\
% %    \text{IfcBuildingElementProxy}(m'') & \text{otherwise}
% %   \end{cases}
% % \end{equation}
The class-mapped object (\ie IFC object) inherits the usage definitions (\ie class attributes), including type and property set, quantity, geometry, and containment use pre-defined by the IFC class, with aligned semantic information that reflects its real-world counterpart in the 3D model. Meanwhile, a mesh and geometric representation for each IFC object is created by parsing the vertices \( V'' \) and faces \( V''_f \). This process determines its placement within the IFC project's 3D space and formulates a geometric representation.

\paragraph{Partition-based Color Coding.} To assign a distinct color representation for each object entity to the BIM model, this step initiates with the retrieval of the vertices color $V''_{rgb}$. Consider the process of calculating the average color of a segment in a 3D model, represented by the equation \( \overline{G}_{rgb} = \frac{1}{|V''_{rgb}|} \sum_{V'' \in m''} V''_{rgb} \). In this context, \( \overline{G}_{rgb} \) symbolizes the average RGB color vector for $m''$. The denominator, \( |V''_{rgb}| \), denotes the number of color vectors within the segment. By calculating \( \overline{G}_{rgb} \), we obtain a consistent color scheme representing the true color of the partitioned point segment and create a material color surface representation of $m''$

\section{Experiments}

\paragraph{Case Studies.} Compared to the work by \cite{yu_building_2019}, which proposed a city-scale BIM and classification framework and selected a model located in Atlantic City as the only case study, we extend our case studies to different scene levels (Our scene level count: 3, \cite{yu_building_2019} scene level count: 1) of point cloud models. These cover indoor scenes (S3DIS: Hallway), exterior architectures (BuildingNet: Residential Church; SRBIM (\ie Our point cloud models semantic segmentation dataset.): Educational Edifice, Urban Residential High-Rise, Commercial Skyscraper, and Public Service Complex), an open landscape  (SRBIM: Military Cemetery), and an urban-scale built environment (SensatUrban: Birmingham Block 0).

\section{Results and Discussion}

\begin{figure}[t]
    \begin{subfigure}{0.23\textwidth}
        \includegraphics[width=\linewidth]{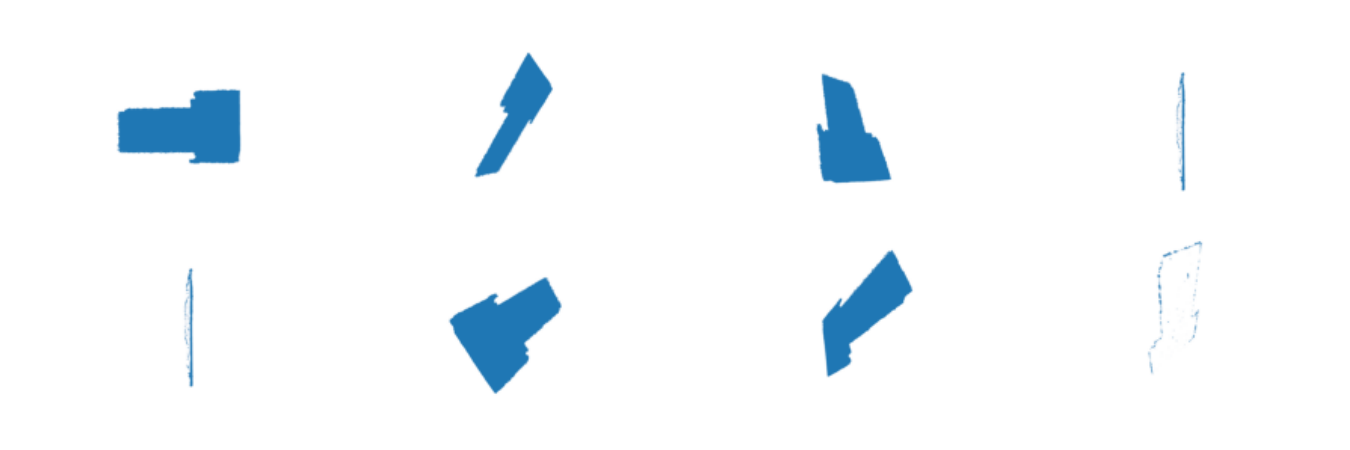}
        % \caption{Ceiling}
        \caption{}
        \label{subfig:Ceiling_unsmoothen}
    \end{subfigure}
    \hfill
    \begin{subfigure}{0.23\textwidth}
        \includegraphics[width=\linewidth]{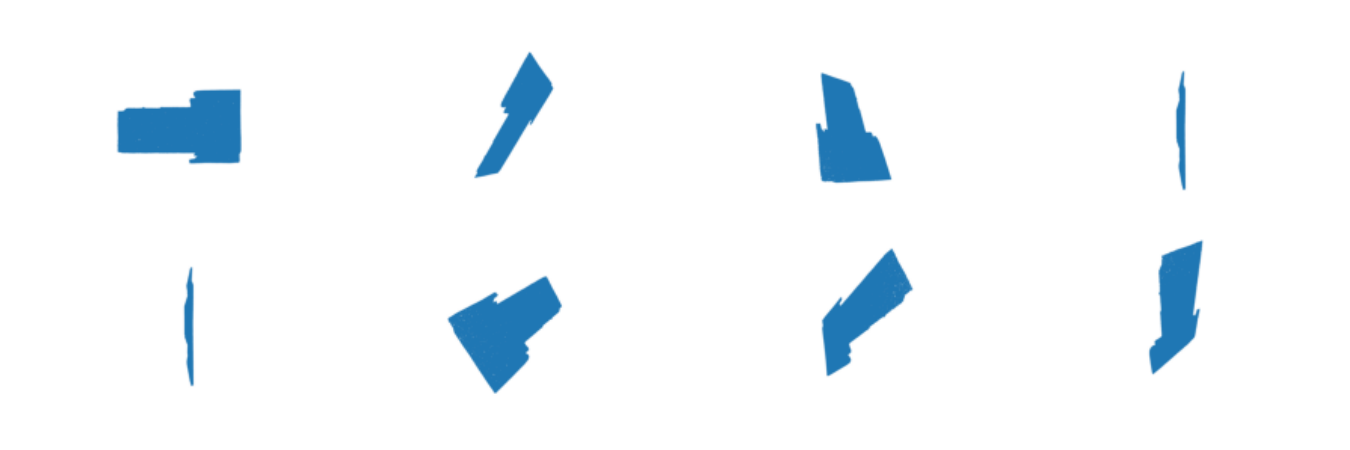}
        % \caption{Ceiling(MFS) }
        \caption{}
         \label{subfig:Ceiling_smoothen}
    \end{subfigure}
    \hfill
        \begin{subfigure}{0.23\textwidth}
        \includegraphics[width=\linewidth]{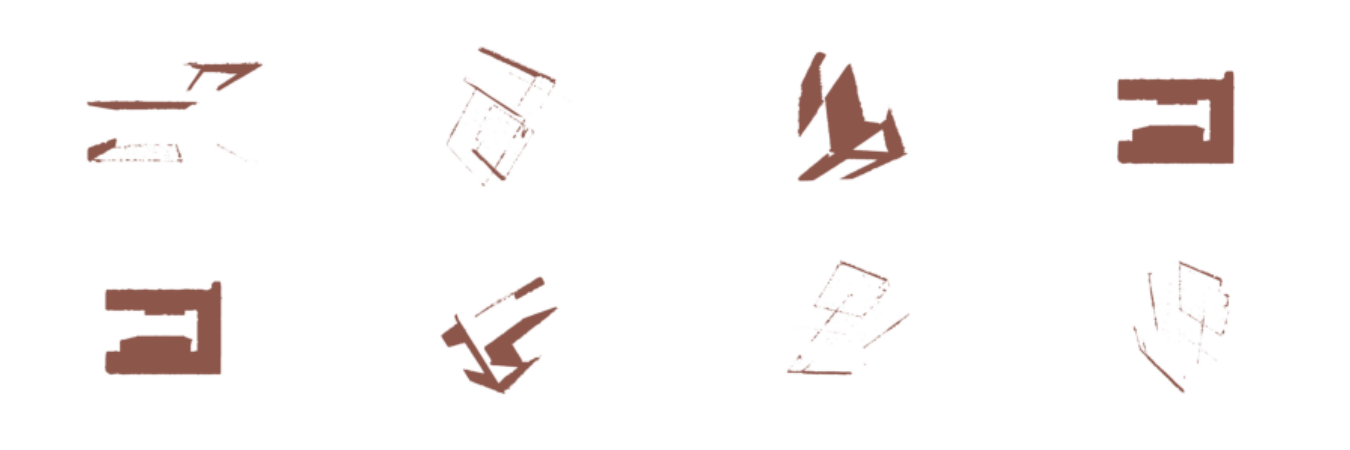}
        % \caption{Wall}
        \caption{}
        \label{subfig:wall_unsmoothen}
    \end{subfigure}
    \hfill
    \begin{subfigure}{0.23\textwidth}
        \includegraphics[width=\linewidth]{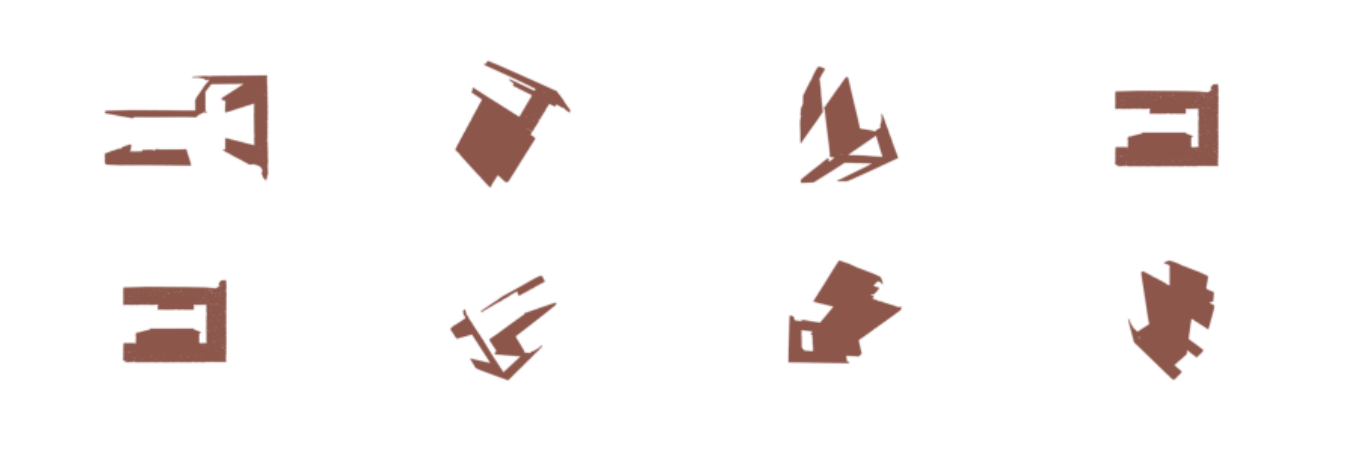}
        % \caption{Wall(MFS)}
        \caption{}
        \label{subfig:wall_smoothen}
    \end{subfigure}
    \caption{Qualitative results of MFS performance on example segments in the S3DIS hallway. Figs (a, c): semantic point cloud segments (\Blue{Blue}: Ceiling, \Brown{Brown}: Wall); Figs (b, d): Refined mesh segments produced by MFS.}
    \label{figs:MFS-abcd}
\end{figure}

\begin{figure}[t]
    \centering
    \includegraphics[width=1\linewidth]{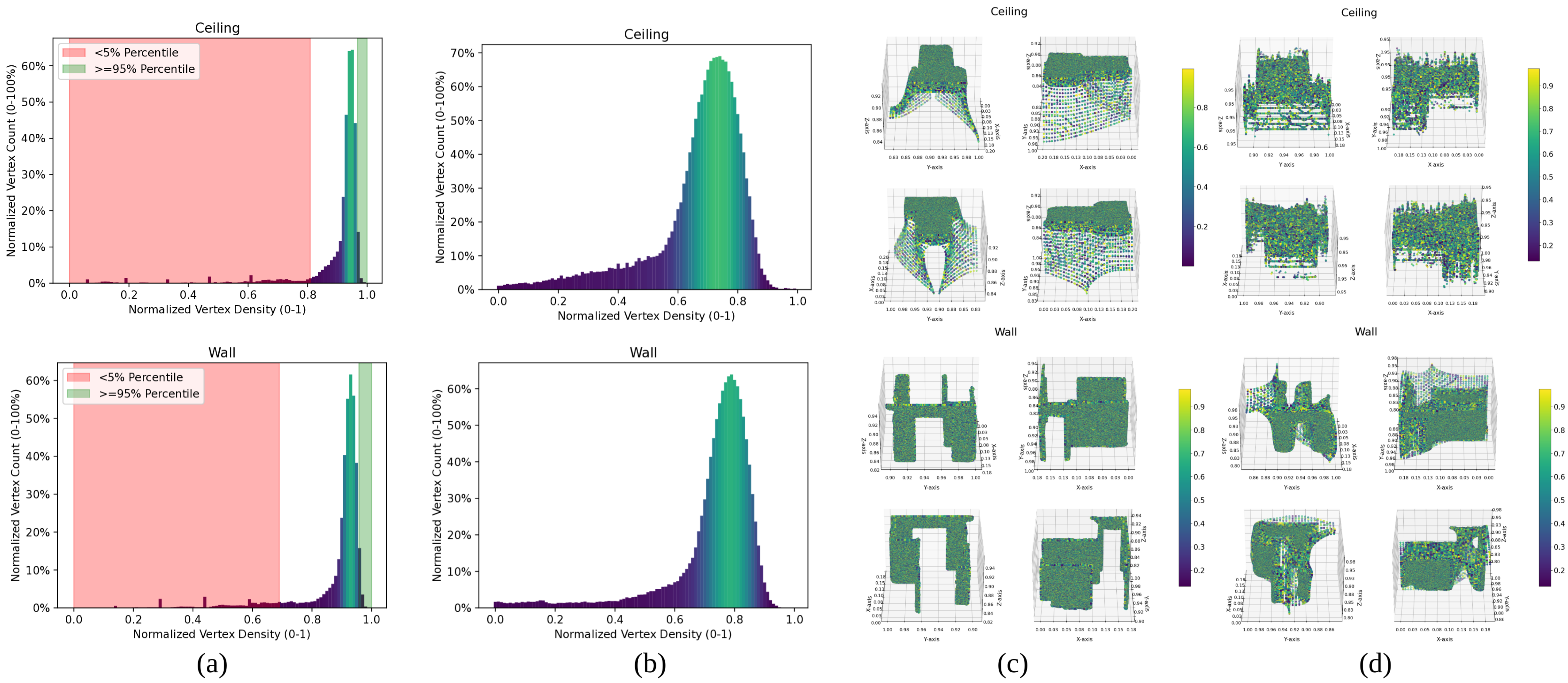}
    \caption{Intermediate results of MFS in the example segments from the S3DIS Hallway. Histograms (a, b) and heatmaps (c, d) compare the vertex density distributions of the mesh segment generated by Poisson Surface Reconstruction before and after post-filtering and smoothing.}
    % \caption{Intermediate results of MFS in the example segments from the S3DIS Hallway.}
    \label{fig:intrem-MFS-S3DIS}
\end{figure}
% \vspace{-\baselineskip}
% \subsubsection{Mesh Reconstruction.}
\paragraph{Mesh Reconstruction.}
Histograms (\cf \cref{fig:intrem-MFS-S3DIS}a and \ref{fig:intrem-MFS-S3DIS}b) reveal the effect of applying a quantile filter to the dataset. In the pre-filter histogram (\cf \cref{fig:intrem-MFS-S3DIS}a), we observe a broader distribution of vertex densities, with a notable presence of vertices in the lower-density regions, as highlighted by the pink area indicating vertices below the $5th$ quantile.

% This suggests a higher level of detail and possible noise or outliers in the mesh, which could be attributed to the inherent variability within the original point cloud data or the reconstruction process itself.

Upon applying the quantile filter, as seen in the post-filter histogram (\cf \cref{fig:intrem-MFS-S3DIS}b), these lower-density vertices are removed. This results in a distribution that is not only more centralized around a pronounced peak but also one that exhibits a reduction in the lower tail variability. The peak, situated near 0.8 on the normalized vertex density scale, reflects a denser and more uniform mesh where the vertices are more consistently distributed.

% The removal of vertices below the $5_{th}$ quantile has ostensibly enhanced the mesh by filtering out sparse regions that may represent noise or non-significant features, leading to a cleaner and potentially more computationally tractable representation. This tightening of the distribution curve is indicative of a mesh that is more reflective of the salient surface features, with less emphasis on the extremities.

Heatmaps (\cf \cref{fig:intrem-MFS-S3DIS}) illustrate the impact of quantile filtering followed by a smoothing operation. Initially, the initial meshes (\cf \cref{fig:intrem-MFS-S3DIS}c) exhibit a wide range of vertex densities, as indicated by the diverse coloration. After the process of quantile filtering and smoothing, the mesh segments (\cf \cref{fig:intrem-MFS-S3DIS}d) show an observable reduction in the lower-density vertices but preserved the denser regions of the initial mesh, with more homogenized vertex density, as evidenced by the more consistent coloration across the mesh surface. The resultant meshes (\cf \cref{figs:MFS-abcd}c and \ref{figs:MFS-abcd}d) are characterized by more regular and visually coherent surfaces for BIM.

\begin{figure}[t]
    \centering
    \includegraphics[width=0.75\linewidth]{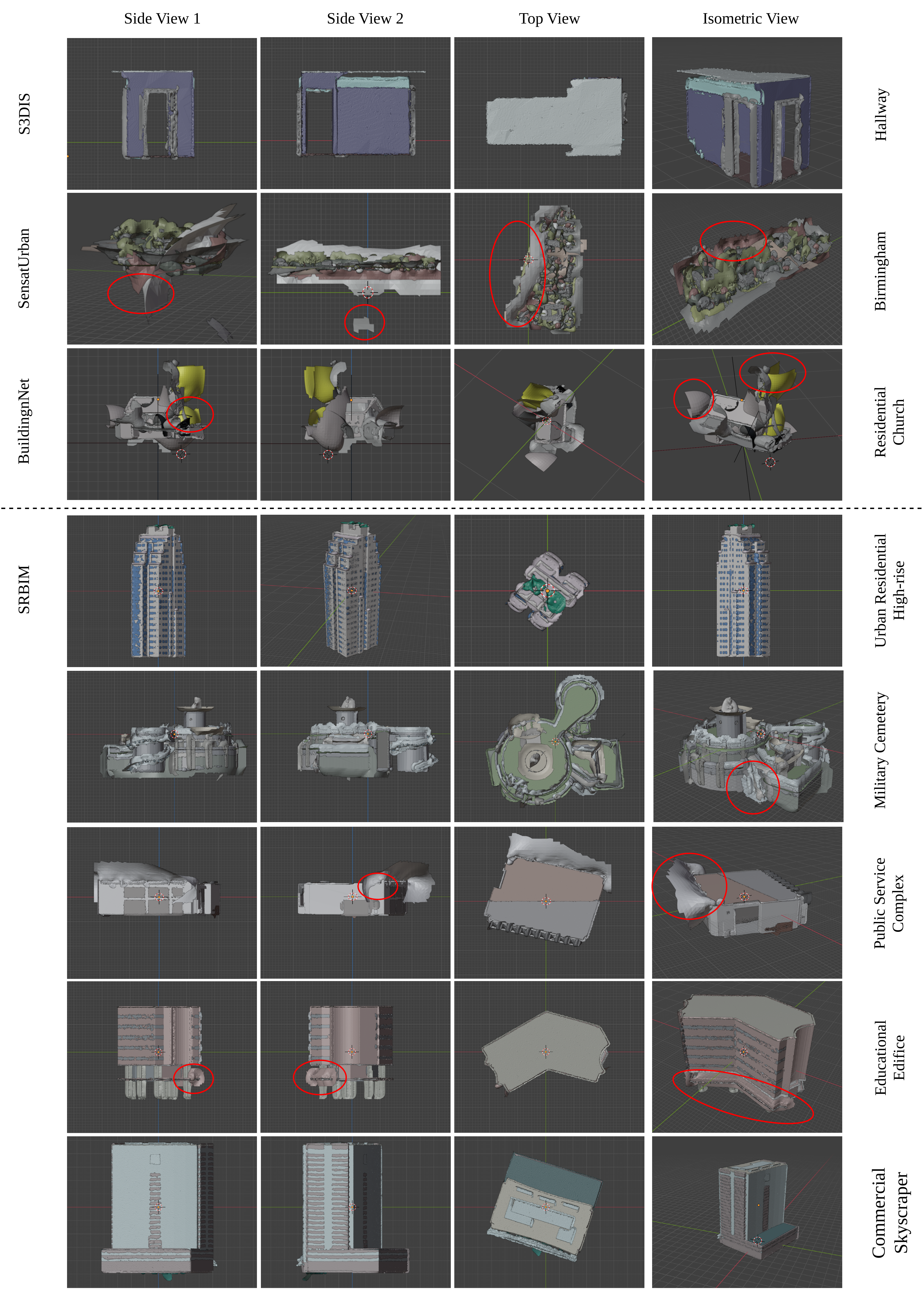}
    \caption{Reconstructed BIM model across datasets: All results are visualized in Blender \cite{blender_blenderorg_2024} with BlenderBIM \cite{blenderbim_blenderbim_2024} add-on. Color-enriched IFC objects are assembled to create a semantic-enriched BIM. The final BIM inherits all attributes and definitions of the IFC objects. IFC objects of distinct classes are instantiated, allowing them to be reused and modified. The \Frst{circled} examples of overextended surfaces indicate areas beyond the intended boundaries of the model structure.}
    \label{Fig:BIM_visual_results}
\end{figure}

\paragraph{BIM Reconstruction.} The final BIM model output results (\cf \cref{Fig:BIM_visual_results}), compared across various scenes, demonstrate SRBIM's ability to generate a high-fidelity model despite a few instances of over-aggregated surfaces. Reconstructed BIM with our SRBIM models and the indoor scene from S3DIS, which showcases the accurate representation of complex scenes. With partition-based color averaging, each built component is distinctively colorized, creating visually appealing BIM. The colorized model aids in easy navigation, management, and error detection of the digital model \cite{dsd_dsd_2019,emsd_emsd_2019,liucci_revit_2023}.

\section{Conclusion}
We introduced SRBIM, a unified framework for semantic BIM modeling. The effectiveness of SRBIM was proven through extensive evaluations, realizing automated BIM modeling for real-world construction projects. 
% Additionally, we also presented the first publicly available 3D building model dataset for BIM reconstruction and 3D semantic segmentation to encourage advancements in fully automated BIM modeling and point cloud segmentation of architectural exteriors.

\newpage

%------------------------------------------------------------------------
% \section{Final copy}

% You must include your signed IEEE copyright release form when you submit your finished paper.
% We MUST have this form before your paper can be published in the proceedings.

% Please direct any questions to the production editor in charge of these proceedings at the IEEE Computer Society Press:
% \url{https://www.computer.org/about/contact}.

%%%%%%%%% REFERENCES
{\small
\bibliographystyle{ieeenat_fullname}
% \bibliography{references_latex}
\bibliography{references}
% \bibliography{egbib}
}
% \bibliographystyle{splncs04}
% \bibliography{egbib}
% \bibliography{references}
\end{document}